\documentclass{scrartcl}
\usepackage[utf8]{inputenc}
\usepackage{amsmath}
\usepackage[english]{babel}
\usepackage[margin=1in]{geometry}
\usepackage{setspace}
\usepackage{natbib}
\usepackage{subfig}
\usepackage{graphicx}
\usepackage{makecell}
\usepackage{multicol}
\usepackage{booktabs}
\usepackage{caption}
\usepackage{subfig}
\usepackage[utf8]{inputenc}
\RequirePackage[table]{xcolor}
\usepackage{subfiles}
\usepackage{multirow}
\usepackage{authblk}

\definecolor{PANDarkGray}{RGB}{153,153,153}

\usepackage{tikz}
\usetikzlibrary{shapes,arrows,positioning}
\usetikzlibrary{shapes.geometric, arrows}
\tikzstyle{process} = [rectangle, rounded corners, minimum width=3cm, minimum height=1cm,
text centered, text width=5cm, draw=black, fill=white]
\tikzstyle{subcat} = [rectangle, minimum width=2cm, minimum height=1cm, 
text centered, text width=2cm, draw=black, fill=white]
\tikzstyle{arrow} = [thick,->,>=stealth]

\usepackage{natbib}

\usepackage{hyperref}

\title{Political DEBATE: Efficient Zero-shot and Few-shot Classifiers for Political Text}
\author[1]{Michael Burnham}
\author[2]{Kayla Kahn}
\author[3]{Ryan Yang Wang}
\author[3]{Rachel X. Peng}
\affil[1]{Department of Politics, Princeton University}
\affil[2]{Department of Political Science, The Pennsylvania State University}
\affil[3]{Manship School of Mass Communication, Louisiana State University}

\date{September 2, 2024}

\begin{document}

\maketitle
\begin{abstract}
\noindent Social scientists quickly adopted large language models due to their ability to annotate documents without supervised training, an ability known as zero-shot learning. However, due to their compute demands, cost, and often proprietary nature, these models are often at odds with replication and open science standards. This paper introduces the Political DEBATE (DeBERTa Algorithm for Textual Entailment) language models for zero-shot and few-shot classification of political documents. These models are not only as good, or better than, state-of-the art large language models at zero and few-shot classification, but are orders of magnitude more efficient and completely open source. By training the models on a simple random sample of 10-25 documents, they can outperform supervised classifiers trained on hundreds or thousands of documents and state-of-the-art generative models with complex, engineered prompts. Additionally, we release the PolNLI dataset used to train these models -- a corpus of over 200,000 political documents with highly accurate labels across over 800 classification tasks.
\end{abstract}

\doublespacing

\section{Introduction}
Text classification is widely used in various applications, such as opinion mining and topic classification \citep{minaee2021deep}. In the past, classification was a technical and labor intensive task requiring a significant amount of manual labeling and a strong understanding of machine learning methods. Recently developed large language models (LLMs), like ChatGPT, have all but eliminated this barrier to entry due to their ability to label documents without any additional training, an ability known as zero-shot classification \citep{ziems2024can, gilardi2023chatgpt, burnham2024stance, rytting2023towards}. Because of this, it is little wonder that LLMs have received widespread adoption within political and other social sciences. 

Yet, despite their convenience, there are strong reasons why researchers should be hesitant to use LLMs for text analysis. The most widely used and performative models are proprietary, closed models. Historical versions of the models are not archived for replication purposes, and the training data is not publicly released. This makes their use at odds with standards of open science. Further, these models have large compute requirements, and charge for their use -- labeling datasets of any significant size can be expensive. We echo the sentiments of \citet{palmer2024using}: Researchers should strive to use open sourced models and should provide compelling justification when using closed models.

We aim to narrow this gap between the advantages of closed, state-of-the-art large language models and the best practices of open science. Accordingly, we present two language models named Political DEBATE (DeBERTa Algorithm for Textual Entailment) Large and Political DEBATE Base. The models are trained specifically for zero and few-shot classification of political text. With only 86 million and 304 million parameters \citep{he2021debertav3}, the DEBATE models are not only a fraction of the size of proprietary models with tens of billions of parameters, such as Claude 3.5 Sonnet \citep{claude35sonnet}, but are as good or better at zero-shot classification of political documents. We further demonstrate that the DEBATE models are few-shot learners without any active learning scheme: A simple random sample of only 10--25 labeled documents is sufficient to teach the models complex labeling tasks when necessary. 

We accomplish this in two ways. First, we use domain specific training with tightly controlled data quality. By focusing the model on a specific domain, the model size necessary for high performance is significantly reduced. Second, we adopt the natural language inference (NLI) classification framework. This allows us to train encoder language models (e.g. BERT \citep{devlin2018bert}) for zero-shot and few-shot classification. These models are much smaller than the generative language models like GPT-4 \citep{openai2023gpt4}. 

Additionally, we release the PolNLI dataset used to train and benchmark the models. The dataset contains over 200,000 political documents with high quality labels from a wide variety of sources across all sub-fields of political science. Finally, in the interests of open science, we commit to versioning both the models and datasets and maintaining historical versions for replication purposes. We outline the details of both the data and the NLI framework in the following sections.

\section{Natural Language Inference: What and Why}
Natural language inference (also known as textual entailment) can be thought of as a universal classification framework. A document of interest, known as the ``premise,'' is paired with a user generated statement, known as the ``hypothesis.'' The hypothesis are analogous to a very simple prompt given to a model like GPT-4 \citep{openai2023gpt4} or Llama-3 \citep{llama3modelcard}. Given a premise and hypothesis pair, an NLI classifier is trained to determine if the hypothesis is true, given the content of the premise. For example, we might pair a tweet from Donald Trump: ``It's freezing and snowing in New York -- we need global warming!'' with the hypothesis ``Donald Trump supports global warming''. The model would then give a true or false classification for the hypothesis -- in this case, true. Because nearly any classification task can be broken down into this structure, a single language model trained for natural language inference can function as a universal classifier and label documents across many dimensions without additional training. 

    

Natural language inference has a number of advantages and disadvantages in comparison to generative LLMs. Perhaps the most significant advantage is that NLI can be done with much smaller language models. While a standard BERT model with 86 million parameters can be trained for NLI, the smallest generative language models capable of accurate zero-shot classification have 7-8 billion parameters \citep[e.g.]{WeiEtal2022FinetuneZeroshot}, and state-of-the-art LLMs have tens to hundreds of billions of parameters \citep{minaee2024large}. In practical terms, this is the difference between a model that can feasibly run on a modern laptop, and one that requires a cluster of high-end GPUs.


The primary tradeoff between NLI classifiers and generative LLMs is between efficiency and flexibility. While an NLI classifier can be much smaller than an LLM, they are not as flexible. LLMs like GPT-4 \citep{openai2023gpt4} and Llama \citep{touvron2023llama} can accept long prompts that detail multiple conditions to be met for a positive classification. In contrast, the hypotheses accepted by an NLI classifier should be short and reduce the task to a relatively simple binary. Many classification tasks are not easily reduced to simple hypothesis statements.

This capability stems from the wide knowledge base about the world that LLMs hold within their weights. Because they are trained on such a massive amount of data, their training distributions contain a wider variety of tasks (e.g. classification, summarizing, programming) and domains (e.g. politics, medicine, history, pop-culture). Such a vast knowledge base requires a much larger model with higher compute demands. Often, much of the knowledge contained in these weights is superfluous to the classification task a researcher may be using them for. Thus, while LLMs  have shown impressive capabilities in zero-shot settings \citep{ziems2024can}, they are inherently very inefficient tools for any \emph{particular} classification task. 

While we acknowledge that generative LLMs can play a valuable role in political research, their necessarily large size and usually proprietary nature also poses a challenge for open science standards. Their compute demands can be expensive, proprietary models are not archived for scientific replication purposes, and the lack of transparency regarding model architectures and training datasets complicates efforts to replicate or improve these models \citep{spirling2023open}. As a result, despite their impressive capabilities and ease of use, the use of proprietary LLMs as a classification tool at least merits explicit justification in a scientific setting \citep{palmer2024using}.

Here, we demonstrate much smaller models can often offer the convenience and performance of generative LLMs by adopting the NLI classification framework and narrowing its domain of expertise from the entire world to the political world. The advantages of our models presented here over LLMs is first, that they are smaller and thus can be more easily trained or deployed on local or free hardware. Second, they are similarly performative to state-of-the-art LLMs on tasks within their domain. Third, they can be easily versioned and archived for reproduciblity. And finally, they are truly open source in that the model architecture and all of its training data is publicly available for scrutiny or future development.

\section{The PolNLI Dataset}
To train our models, we compiled the PolNLI dataset -- a corpus of 201,691 documents and 852 unique entailment hypothesis. We group these hypotheses into four tasks: stance detection (or opinion classification), topic classification, hate-speech and toxicity detection, and event extraction. Table \ref{tab:datasets} presents the number of datasets, unique hypotheses, and documents that were collected for each task. PolNLI a wide variety of sources including social media, news articles, congressional newsletters, legislation, crowd-sourced responses, and more. We also adapted several widely used academic datasets such as the Supreme Court Database \citep{CiteSupremeCourtDB} by attaching case summaries to the dataset's topic labels. The vast majority of text included in PolNLI is human generated --- only a single dataset containing 1,363 documents is generated by an LLM.

\begin{table}[h]
\centering
\begin{tabular}{|l|c|c|c|}
\hline
\textbf{Task} & \textbf{Datasets} & \textbf{Hypotheses} & \textbf{Documents} \\
\hline
Stance Detection & 11 & 361 & 66,581 \\
\hline
Topic Classification & 5 & 278 & 62,005 \\
\hline
Hate-Speech/Toxicity & 2 & 177 & 41,871 \\
\hline
Event Extraction & 4 & 36 & 31,234\\
\hline\hline
Total & 22 & 852 & 201,691 \\
\hline
\end{tabular}
\caption{Summary of Tasks, Datasets, Hypotheses, and Documents}
\label{tab:datasets}
\end{table}

In constructing the PolNLI dataset, we  prioritized both the quality of the labels and the diversity of the data sources. We used a five step process to accomplish this:
\begin{enumerate}
    \item Collecting and vetting datasets.
    \item Cleaning and preparing data.
    \item Validating labels.
    \item Hypothesis Augmentation.
    \item Splitting the data.
\end{enumerate}

\subsection{Collecting and Vetting Datasets}
We identified a total of 48 potential datasets from replication archives, the HuggingFace hub, academic projects, and government documents. A complete list of datasets we used is located in Appendix \ref{appendix:sources}. Several of the collected datasets had been compiled by their authors for other classification tasks while others --- like the Global Terrorism Database \citep{CiteGTD} and the Supreme Court Database \citep{CiteSupremeCourtDB} --- were adapted from general purpose public datasets. We also compiled several new datasets specifically for this project in order to address gaps in the training data. For each dataset, we reviewed the scope of the data, the collection and labeling process, and made a qualitative assessment of the data quality. Datasets for which we determined the quality of the data to be too low or redundant with sources already collected were omitted. 

\subsection{Cleaning and Preparing Data}
To clean the data, we took care to remove any superfluous information from documents that the models might learn to associate with a particular label. This includes aspects like news outlet identifiers in the headings of articles or event records that start each entry with a date. No edits were made to document formatting, capitalization, or punctuation in order to maintain variety in the training data.


For each unique label in the data, we manually created a hypothesis that correlated with that label. For example, documents that were labeled for topic or event were paired with the hypothesis ``This text is about (topic/event type)'' and documents labeled for stance were paired with the hypothesis ``The author of this document supports (stance).'' Most hypotheses are framed as descriptive statements about the document, as in the two previous examples.

Finally, each document-hypothesis pair was assigned an entail/not entail label based on the label from the original dataset. For example, a document labeled as an expression of concern over global warming would be paired with the hypothesis ``The author of this text believes climate change is a serious concern'' and be assigned the ``entail'' label.\footnote{While several other NLI datasets, such as SNLI, have adopted an entail, neutral, contradict labeling scheme, we opted for the simpler entail/not entail because it was a common scheme that all of the collected datasets could be adapted to. Accordingly, neutral and contradiction labels were combined into the ``not entail'' label.}

One challenge with this approach is that topic and event data only contained positive entailment labels. That is, if an event summary was about a terrorist attack, it was with the hypothesis ``This document is about a terrorist attack'' and the entailment labels for these were initially always true. However, we wanted to train the model to not only recognize what is a terrorist attack, but what is not a terrorist attack. To accomplish this for datasets and documents that needed negative cases, we duplicated the documents and then randomly assigned one of the other topic or event hypotheses, and then assigned a ``not entail'' label. One concern is that documents can contain multiple topics, and might be assigned a topic they are related to by chance. This concern is addressed through the validation process outlined in the next section.

\subsection{Validating Labels}
The original curators of the collected datasets used many approaches to labeling their data with varying levels of rigor. The accuracy of labels is critically important to training and validating models, and thus we wanted to ensure that only high quality labels were retained in our data. To meet this objective, we leveraged the much larger language models, GPT-4 and GPT-4o. Recent research has shown that LLMs are as good, or better, than human coders for similar classification tasks \citep{burnham2024stance, chang2024survey, gilardi2023chatgpt}. We thus used these proprietary LLMs to reclassify each collected document with a prompt containing an explanation of the task and the entailment hypotheses we generated. A template for the prompt is contained in Appendix \ref{appendix: prompts}. We then removed documents where the human labelers and the LLM disagreed. To ensure that the LLMs were generating high quality labels, we took a random sample of 400 documents labeled by GPT-4o and manually reviewed the labels again. We agreed with the GPT-4o labels 92.5\% of the time, with a Cohen's $\kappa$ of 0.85. Of the 30 documents where there was disagreement, 16 were judged to be reasonable disagreements where the document could be interpreted either way. The remaining 14, or 3.5\% of all documents, were labeled incorrectly by the LLM.

\subsection{Hypothesis Augmentation}
An ideal NLI classifier will produce identical labels if a document is paired with different, but synonymous, hypotheses (e.g. the hypotheses ``This document is about Trump'' and ``This text discusses Trump'' should yield similar classifications). To make our model more robust to the various phrasings researchers might use for hypotheses, we presented each hypothesis to GPT-4o and then asked it to write three synonymous sentences. We then manually reviewed the LLM generated hypotheses and removed any that we felt were not sufficiently similar in meaning. Each document was then randomly assigned an ``augmented hypothesis'' from a set containing the original hypothesis and the generated alternatives. Finally, we manually varied hypotheses by randomly substituting a few very common words with synonymous words (e.g. text/document, supports/endorses). In total, this increased the number of unique entailment phrases to 2,834.

\subsection{Splitting the Data}
To split the data into training, validation, and test sets we proportionally sampled from each of the four tasks to construct testing and validation sets of roughly 15,000 documents each. The rest of the data were allocated to the training set. Because we wanted to evaluate model performance in a zero-shot context, a simple random sampling approach to splitting the data would not work. Instead, we randomly sampled from the set of unique hypotheses and allocated all documents with those hypotheses to the test set. This ensures that models did not see any of the test set hypotheses, or their synonymous AI generated variants, during training. The validation set consists of roughly 10,000 documents with hypotheses that are not in the training set, and 5,000 documents with hypotheses that are in the training set. This allows us to both estimate the model's zero-shot performance during testing, as well as look for evidence of over-fitting if performance diverges between the hypotheses seen and not seen during training.

\section{Training}
The foundation models we used for training were a pair of DeBERTa V3 base and large models fine tuned for general purpose NLI classification by \citet{laurer_building_2023}. We use these models for a number of reasons: First, the DeBERTa V3 architecture is the most performative on NLI tasks among transformer language models of this size \citep{wang2019superglue}. Second, using models already trained for general purpose NLI classification allows us to more efficiently leverage transfer learning. Before we used these models for our application, they were trained on five large datasets for NLI, and 28 smaller text classification datasets. This means that we begin training with a model that already understands the NLI framework and general classification tasks, allowing it to more quickly adapt to the specific task of classifying political texts \citep{laurer2022annotating}. 

We used the Transformers library \citep{wolf2020transformers} to train the model and monitored training progress with the Weights and Biases library \citep{wandb}. After each training epoch (an entire pass through of the training data), model performance was evaluated on the validation set and a checkpoint of the model was saved. We selected the best model from these checkpoints using both quantitative and qualitative approaches. The model's training loss, validation loss, Matthew's Correlation Coefficient (MCC), F1, and accuracy was reported for each checkpoint. We then tested the best performing models according to these metrics by examining performance on the validation set for each of the four classification tasks, and across each of the datasets. This helped us to identify models with consistent performance across task and document type. 

Finally, we qualitatively assessed the models by examining their behavior on individual documents. This included introducing minor edits or re-phrasings of the documents or hypotheses so that we could identify models with stable performance that were less sensitive to arbitrary changes to features like punctuation, capitalization, or synonymous word choice. Hyperparameters used to train the models are in Appendix \ref{appendix: params}.

\section{Zero-shot Learning Performance}
We benchmark our models on the PolNLI test set against four other models that represent a range of options for zero-shot classification. The first two models are the DeBERTa base and DeBERTa large general purpose NLI classifiers trained by \citet{laurer_building_2023}. These are currently the best NLI classifiers that are publicly available \citep{laurer_building_2023}. We also test the performance of Llama 3.1 8B, an open source generative LLM released by Meta \citep{llama3modelcard}. This model is the smallest version of Llama 3.1 released and represents a generative LLM that can feasibly be run on a desktop computer with a high end GPU, or a CPU with an integrated GPU like the Apple M series chips in modern macbooks. Finally, we benchmark Calude 3.5 Sonnet \citep{claude35sonnet}. This model is a state-of-the-art proprietary LLM. At the time of writing, it is widely considered to be among the best models available \citep{syed2024benchmarkllm}. Notably, we do not include GPT-4o in our benchmark because it was used in the validation process from which the final labels were derived. We discourage bench-marking OpenAI models on the PolNLI dataset for this reason.

We use MCC as our primary performance metric due to its relative robustness to other metrics like F1 and accuracy on binary classification tasks \citep{chicco2020advantages, chicco2023matthews}. MCC is a special case of the Pearson correlation coefficient and can be interpreted similarly. It rangers from -1 to 1 with higher values indicating greater performance. 

\subsection{PolNLI Test Set}
\begin{figure}
    \centering
    \includegraphics[width=\linewidth]{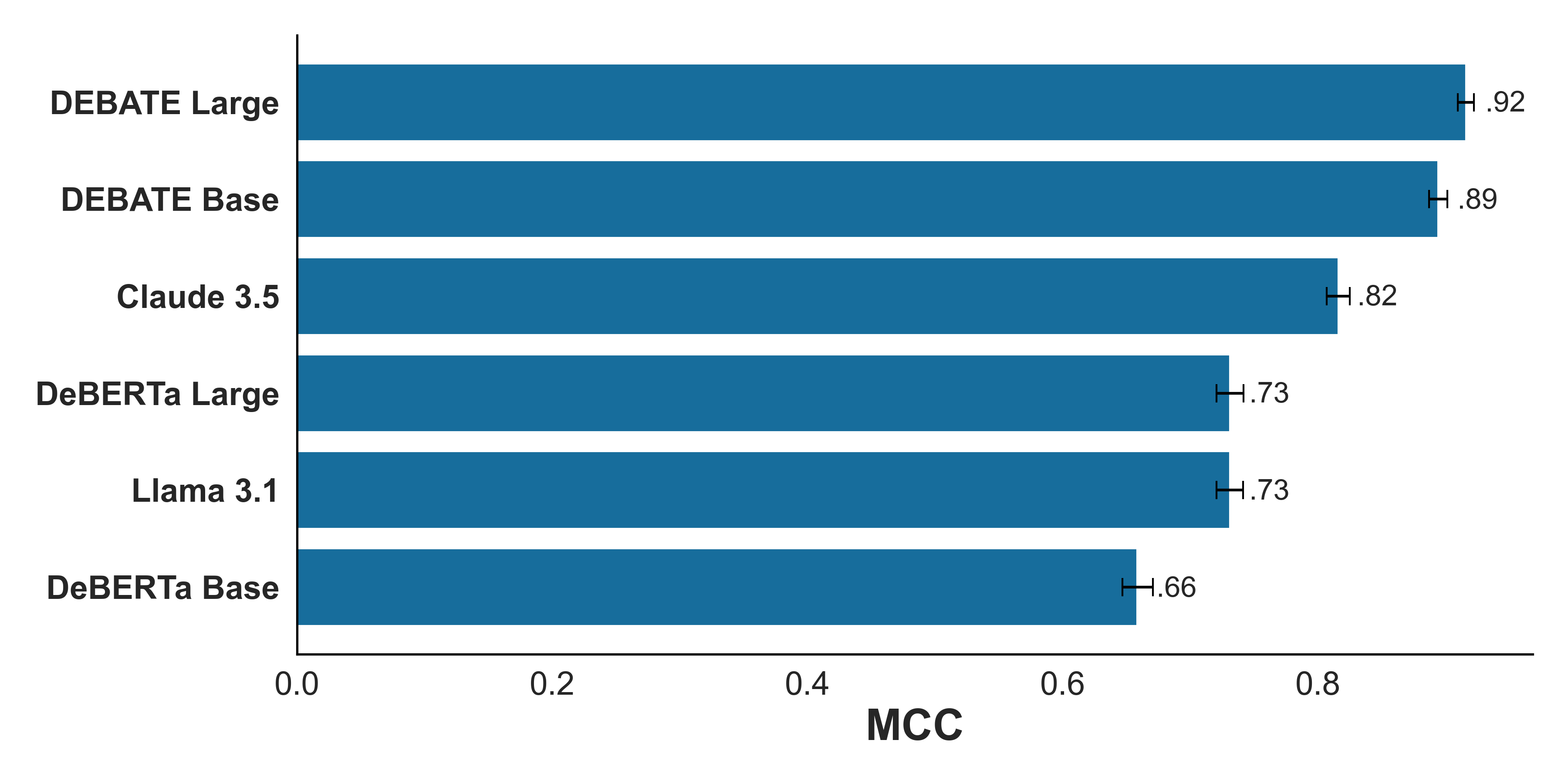}
    \caption{Zero-shot performance of all four tasks for each model}
    \label{fig:test_overall}
\end{figure}

Figure \ref{fig:test_overall} plots performance with bootstrapped standard errors across all four tasks for each model. We observe that the DEBATE models are more performative than alternatives when all tasks and datasets are combined.

\begin{figure}
    \centering
    \includegraphics[width=\linewidth]{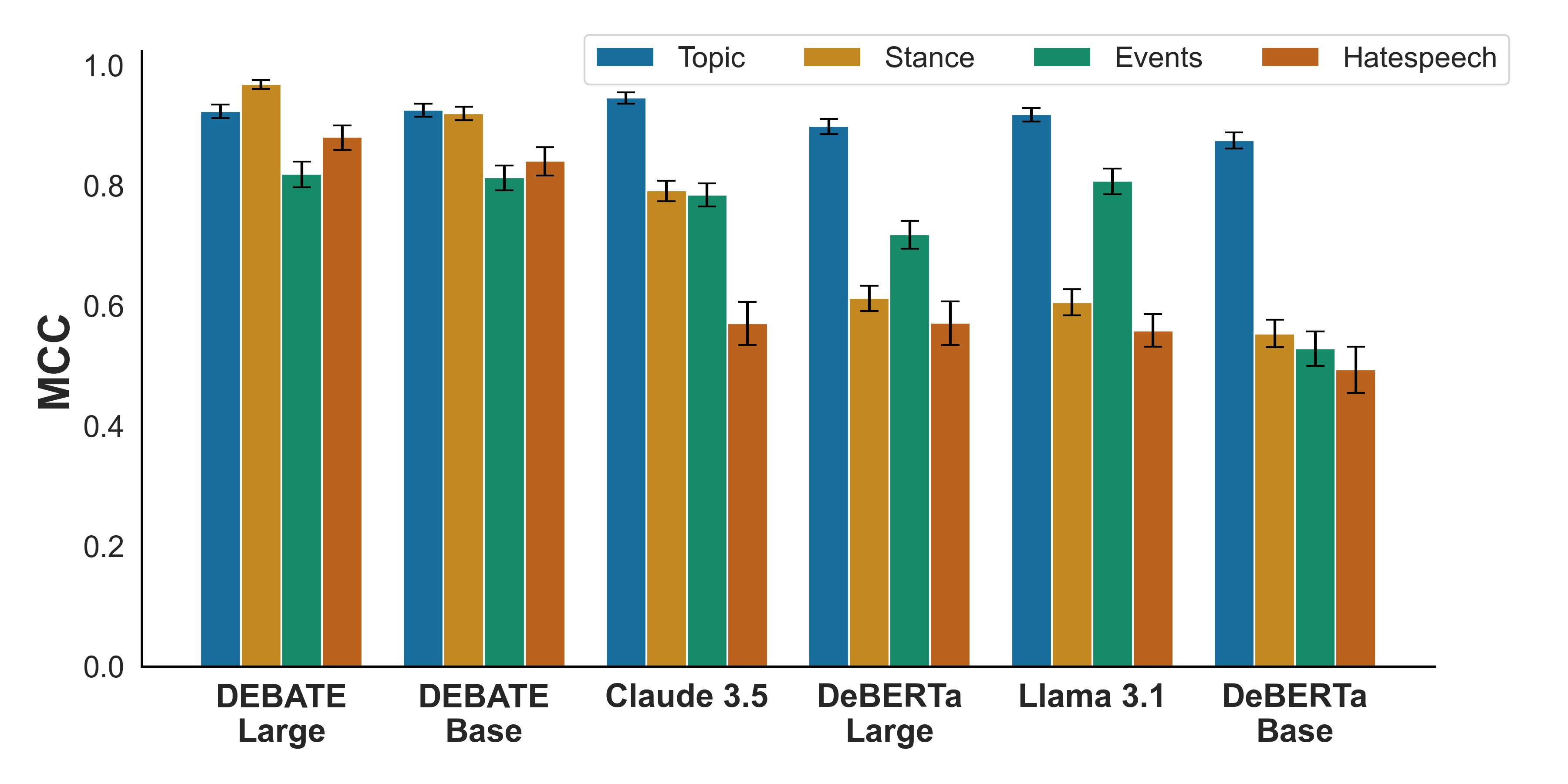}
    \caption{Zero-shot performance of each task for each model}
    \label{fig:test_tasks}
\end{figure}

In figure \ref{fig:test_tasks} we break out performance across our four tasks: Topic classification, stance detection, event extraction, and hate-speech identification. While all models perform well on topic classification, significant gaps emerge on the other tasks. The DEBATE models and Claude 3.5 Sonnet perform significantly better than the other models on stance detection. On event extraction tasks we see comparable performance between the DEBATE models and the two generative LLMs, Claude 3.5 and Llama 3.1. Perhaps the most notable gap in performance is on the hate-speech detection task -- the DEBATE models perform significantly better than the other models. We think that this is likely because hate-speech is a highly subjective concept and our models are better tuned to the particular definitions used in the datasets we collected.

\begin{figure}
    \centering
    \includegraphics[width=\linewidth]{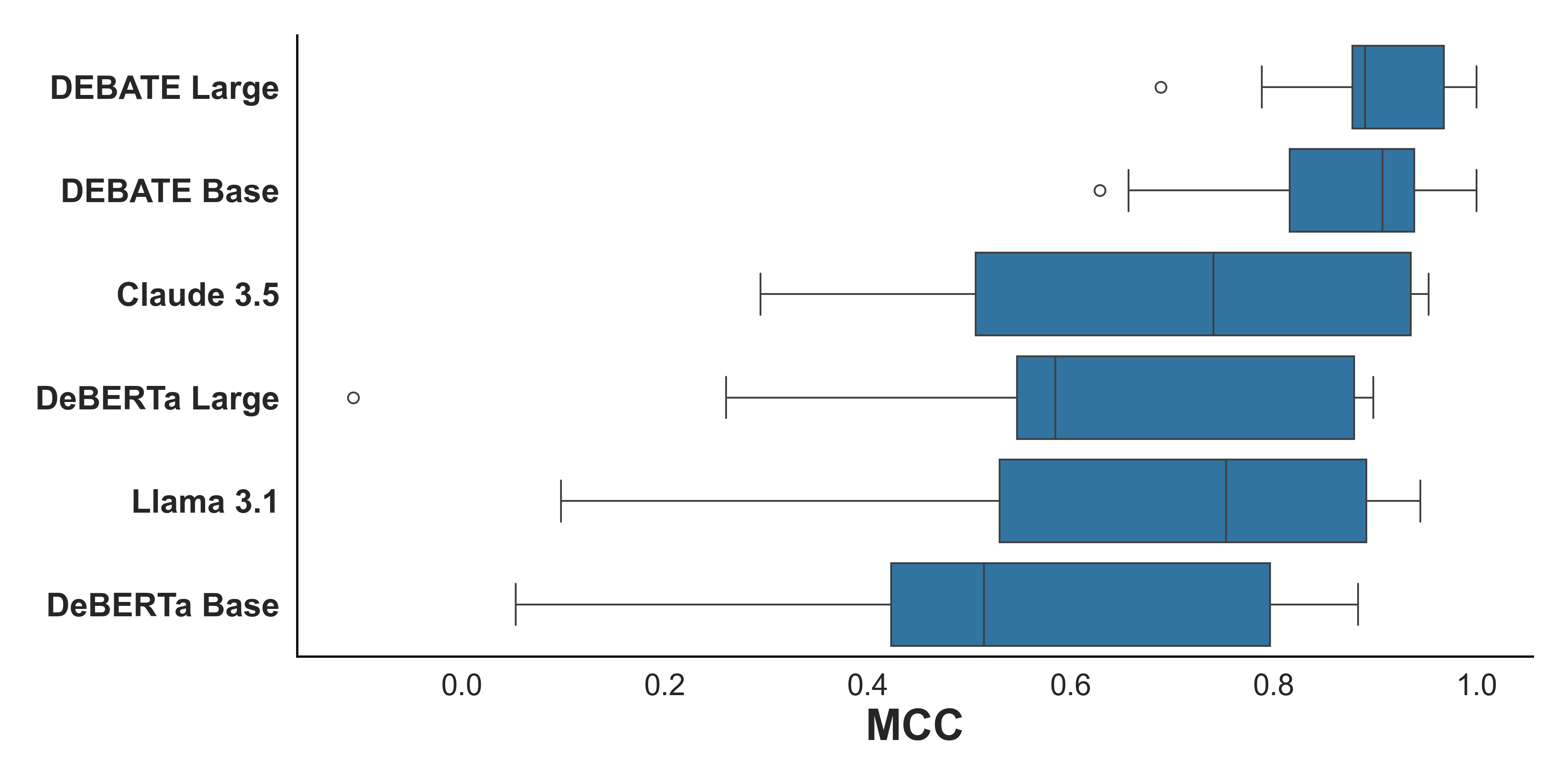}
    \caption{Zero-shot learning MCC distribution of four tasks for each model}
    \label{fig:test_datasets}
\end{figure}

Finally, in figure \ref{fig:test_datasets}, we plot this distribution of performance across all datasets in the test set. We again observe that the DEBATE models are more consistently performative than alternatives. For most models, the Polistance Quote Tweets dataset was the most challenging dataset, with the DeBERTa Large model having a negative correlation with the correct classification. This dataset measures stance detection and is particularly challenging for language models to parse because quote tweets often contain two opinions from two different people. The model has to parse both of these opinions and correctly attribute stances to the right authors. Even the state-of-the-art Claude 3.5 had an MCC of only 0.29 on the task. However, because the Political DEBATE models were explicitly trained to parse such documents, the base and large models were able to achieve MCCs of 0.62 and 0.88 respectively.

\section{Few-shot Learning Performance}
One advantage of the NLI classification framework is that models trained for NLI can more quickly adapt to other classification tasks \citep{laurer2022annotating}. Few-shot learning refers to the ability to learn a new classification task with only a few examples. Whereas a conventional supervised classifier usually requires hundreds or even thousands of labeled documents to train, models like GPT-4 and Claude 3.5 have demonstrated the ability to improve classification with only a handful of examples provided in the prompt.

Here, we demonstrate that domain adapted NLI classifiers are efficient few-shot learners. With a random sample of 10--25 documents and no active learning scheme, these models can learn new classification tasks at levels comparable to, or better than, supervised classifiers and generative language models. We use two examples from other research projects to illustrate this capability. The first comes from the Mood of the Nation poll, a regular poll issued by the McCourtney Institute for Democracy which recently begun using Llama 3.1 as part of its annotation process for open-text survey questions \citep{berkman2024democracy}. The second is from \citet{block2022perceived} who trained a transformer model on roughly 2,000 tweets to identify posts that minimize the threat of COVID-19.

For our testing procedure we first use both DEBATE models and a simple hypothesis for zero-shot classification on each document. We then take four simple random samples of 10, 25, 50, and 100 documents, train each of the two DEBATE models on these random samples, and then estimate performance of both models for the respective sample size on the rest of the documents. We repeat this 10 times for each training sample size and calculate a 95\% confidence interval. Importantly, we did not search for the best performing hypothesis statements or model hyper-parameters. We simply used the default learning rate and then trained the model for 5 epochs. We felt this was important because a few-shot application assumes researchers do not have a large sample of labeled data to search for the best performing parameters. Rather, few-shot learning should work out-of-the-box to be useful. We also note that that while training these language models on large data sets like PolNLI can take hours or days with a high end GPU, training time in a few shot context is reduced down to seconds or minutes and can be done without high-end computing hardware.

\subsection{Mood of the Nation: Liberty and Rights}
One of the questions on the Mood of the Nation poll is an open text form asking, ``What does democracy mean to you?'' The administrators of this poll had a team of research assistants manually label answers to this question that were related to ``liberty and rights.'' This category was broadly defined as responses that discuss freedoms and rights generally, or specific rights such as speech, religion, or the contents of the bill of rights. However, if the document was exclusively about voting rights, it was assigned to another category. If it mentioned voting rights in addition to other rights, it was still classified as ``liberty and rights.'' This classification task is somewhat difficult to reduce down to a simple hypothesis statement, and is thus a good candidate for few-shot training.

\citet{berkman2024democracy} wanted to automate the labeling of short answer responses and wanted to use open source models to do so. This was motivated both by open science standards, and privacy concerns over uploading responses to a proprietary API like GPT-4o. Llama 3.1 worked well, and classified documents with an MCC of 0.74 and accuracy of 88\%. Discrepancies between the LLM and human coders were judged to be primarily reasonable disagreements.

In a zero-shot context, Llama 3.1 comfortably out performs our models due to its ability to accept prompts with more detailed instructions. However, after only 10 training samples we see a large jump in performance with both the large and base DEBATE models, with Llama not significantly different than either. At 25 documents the large model significantly outperforms llama 3.1, and at 50 documents the base model does as well.

\begin{figure}%
    \centering
    \subfloat[\centering Performance on Mood of the Nation dataset (compared with Llama 3.1)]{{\includegraphics[width=.46\textwidth]{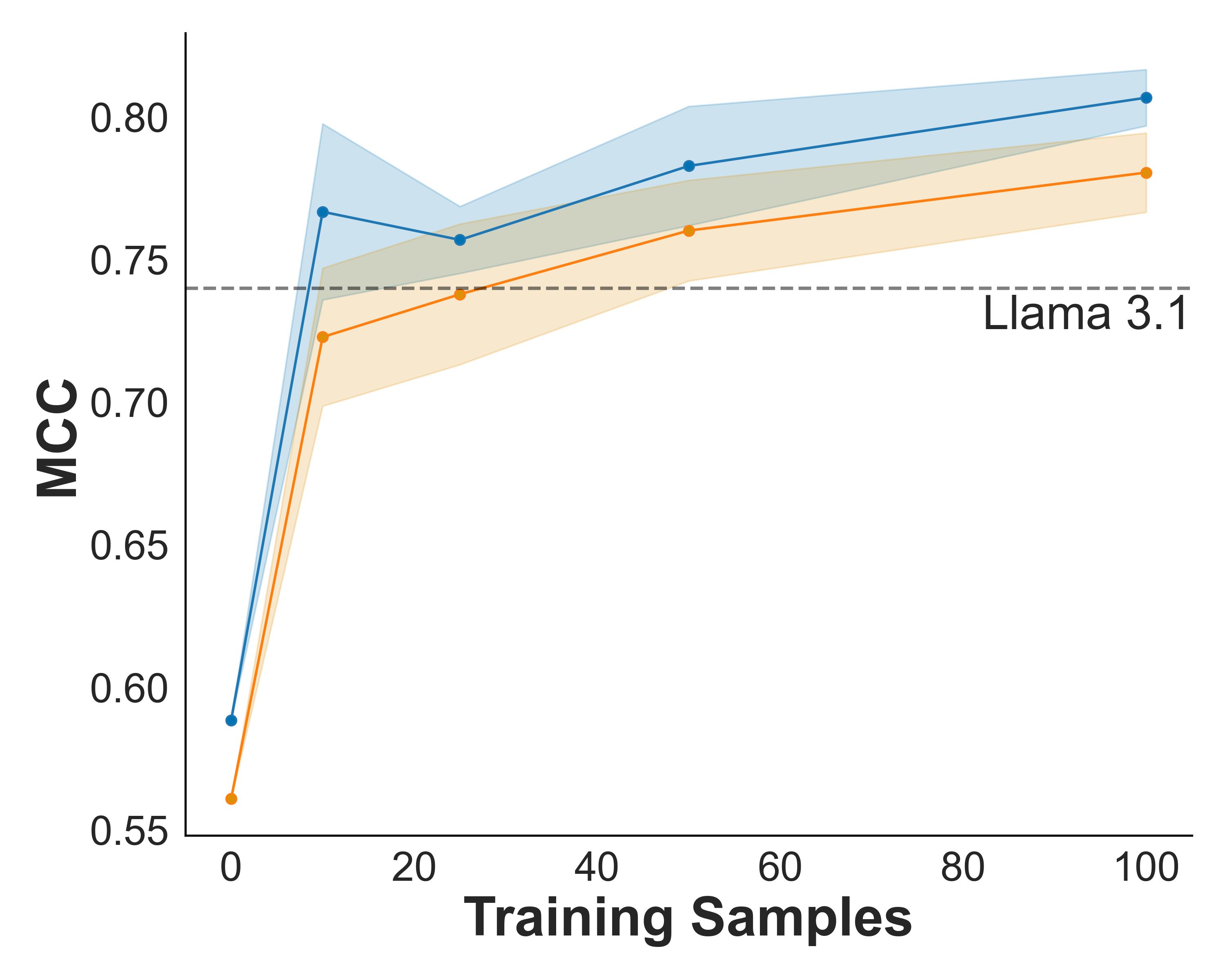} }}%
    \subfloat[\centering Performance on COVID-19 dataset (compared with Electra transformer)]{{\includegraphics[width=.46\textwidth]{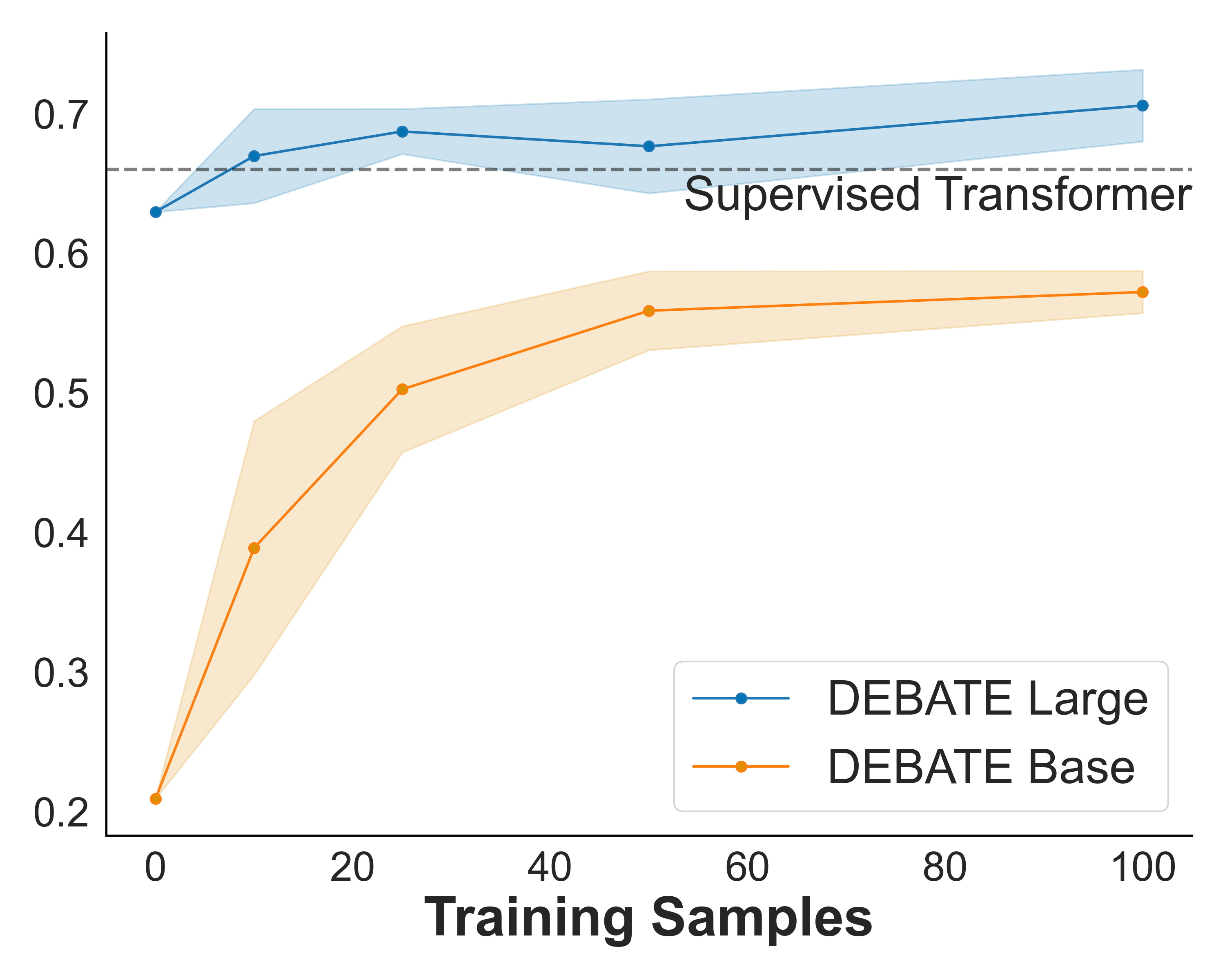} }}%
    \caption{Few-shot learning performance of DEBATE models}%
    \label{fig:example}%
\end{figure}

\subsection{COVID-19 Threat Minimization}
\citet{block2022perceived} classified Twitter posts about COVID-19 based on whether or not they minimized the threat of COVID-19. Threat minimization was defined as anti-vaccination or anti-masking rhetoric, comparisons to the flu, statements against stay-at-home orders, claims that COVID-19 death counts were faked, or general rhetoric that the disease did not pose a significant health threat. This presents a particularly difficult classification challenge because threat minimization of COVID-19 is a somewhat abstract concept and can be expressed in many different ways across disparate topics. To address this, \citet{block2022perceived} trained an Electra transformer on 2,000 tweets with a Bayesian sweep of the hyper-parameter space. This process involved training 30 iterations of the model to find the best performing hyper-parameters. The final model achieved an MCC of 0.66.

For an NLI classifier, the above classification criteria are too numerous to elegantly fit into a single entailment hypothesis. While \citet{burnham2024stance} demonstrated that such tasks can be done zero-shot by dividing it into smaller tasks (e.g. classify the documents once for anti-vaccination rhetoric, another time flu comparisons, and so forth), few-shot learning provides a more elegant solution. To test the models, we use the basic hypothesis ``The author of this tweet does not believe COVID is dangerous.'' Here, we observe that the base model largely fails at the task in a zero-shot context, and fails to match the accuracy of the supervised classifier with 100 training samples. The large model proves more capable of learning the task, matching the supervised classifier at 10 training samples (MCC = 0.67, accuracy = 87\%), and exceeding it at only 25 training samples (MCC = 0.69, accuracy = 88\%).

\section{Timing Benchmarks}
To assess cost-effectiveness, we ran our two DEBATE models and Llama 3.1 across a diverse range of hardware.\footnote{We exclude the DeBERTa models used in the performance bench-marking above because they have the same architecture as the DEBATE models, and thus label documents at the same speed. We also do not time proprietary LLMs because their speed is highly determined by server traffic and we cannot test them in a controlled setting on common hardware. Classification speed is also of more concern when using local hardware because it occupies computing resources that might be need during the run time.} We did so with a random sample of 5,000 documents from the PolNLI test set and the simple hypothesis ``This text is about politics." We selected four different types of hardware. First, the NVIDIA GeForce RTX 3090 GPU provides high-performance, consumer-grade machine learning capabilities, making it a suitable choice for intensive computational tasks. Second, the NVIDIA Tesla T4 is a free GPU available through Google Colab. In contrast to the RTX 3090, the T4 is easy for researchers to access free of charge. Third, we used a Macbook Pro with the M3 max chip. This is a common laptop with a built-in GPU that is integrated in the system-on-chip, as opposed to the RTX 3090 and Tesla T4 which are discrete GPUs. Finally, the AMD Ryzen 9 5900x CPU was utilized to evaluate performance on a general purpose CPU.\footnote{We do not test Llama 3.1 on the Tesla T4 GPU or the Ryzen 5900x CPU. The model is too large to run on the Tesla T4, and slow enough on on a CPU that it's not recommended to do so in any context.}

                                     

\begin{figure}
    \centering
    \includegraphics[width=\linewidth]{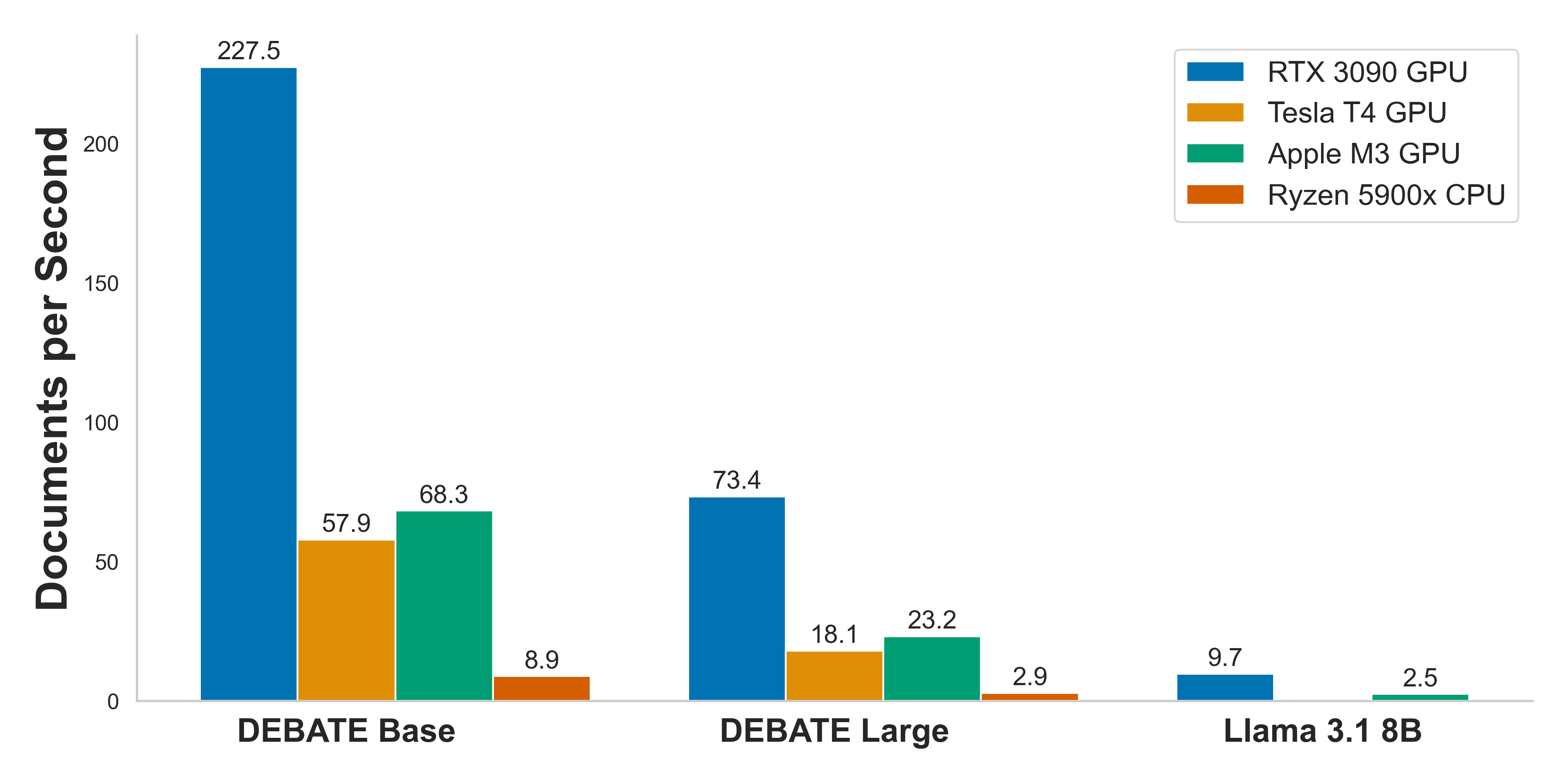}
    \caption{The DEBATE models offer a massive efficiency advantage over generative language models.}
    \label{Timing benchmark of each hardware selection}
\end{figure}

We observe that the DEBATE models offer massive speed advantages over even small generative LLMs like Llama 3.1 8B. While discrete GPUs like the RTX 3090 do offer a large performance advantage, Documents can still be classified at a relatively brisk pace with a laptop GPU like on the M3, or a free cloud GPU like the T4.

\section{Limitations and Model Use}
The \href{https://huggingface.co/mlburnham or by searching for Pol\_NLI}{model and dataset} can be downloaded for free on the \href{https://huggingface.co/mlburnham}{HuggingFace hub}. We recommend using Python's Transformers \citep{wolf2020transformers} and Datasets \citep{Lhoest2021} libraries to use the models and data. In most cases, models and data can be deployed with only a few lines of code. We include boilerplate code for both zero-shot and few-shot applications on the github repository for this paper. While we offer brief advice on application here, for a more thorough exploration of best practices when using NLI classifiers we defer to \citet{burnham2024stance}.


\subsection{Which Model Should I Use?}
We offer the following two guidelines for selecting a model:
\begin{itemize}
    \item Use the large model for zero-shot classification.
    \item Use the large model for most few-shot applications.
    \item Use the base model for simple few-shot tasks or supervised classification.
\end{itemize}

\noindent Both our extensive use of NLI models and previous research \citep{burnham2024stance} indicates that larger models are much better at generalizing to unseen tasks. However, for tasks that are more explicitly within the training distribution such as hate-speech detection or approval of politicians, we expect comparable performance between the large and base models, with the base model offering a significant advantage in efficiency. In the few-shot context we expect similar performance between the large and base model given the results above. However, We also observed that the large model more quickly learns tasks. There is also no clear measurement of a task's simplicity, only qualitative judgements. Thus, we recommend using the large model whenever feasible.

\subsection{When Should I Consider Few-shot or Supervised Training?}
An NLI classifier will be most performative in a zero-shot context under the following conditions:
\begin{itemize}
    \item Labels can reasonably be derived from only the text of a document and do not require meta-knowledge about the document such as who wrote it, when, and under what circumstances.
    \item Labels are for concepts that are commonly understood (e.g. support/opposition to a person or policy) rather than bespoke concepts for a particular research project, or require specialized domain knowledge (e.g. documents about political rights except the right to vote, ``threat minimization'' of COVID-19). 
    \item Documents are short, generally a sentence or paragraph, or can be segmented into short documents.
\end{itemize}
If any of these conditions are not met, you should consider few-shot or supervised training. Whether or not these conditions are met is a qualitative judgment that should be made based on familiarity with the data and task. As with any classification task, you should always validate your results with some manually labeled data.

\subsection{How Should I Construct Hypotheses?}
We recommend using short, simple hypotheses similar to the templates used in the training data. Fore example:
\begin{itemize}
    \item ``This text is about (topic or event)''
    \item ``The author of this text supports (politician or policy position)''
    \item ``This text is attacking (person or group)''
    \item ``This document is hate-speech''
\end{itemize}
While researchers can certainly deviate from these templates, few-shot training may be appropriate for tasks that require long hypotheses with multiple conditions.

\subsection{Other Limitations}
Despite the impressive results demonstrated here, we want to emphasize that researchers should not expect the DEBATE models to outperform proprietary LLMs on all classification tasks. The massive size and training sets of proprietary models inevitably means a larger variety of tasks are in their training distribution. Accordingly, we expect that LLMs will more robustly generalize in the zero-shot context for tasks that are less proximate to what is contained in the PolNLI data set. We recommend few-shot training for such tasks.

We also note that these models are trained exclusively for English documents, and it is unknown how the models would perform if re-trained for non-English documents.

\section{Conclusion and Future Work}
The presented zero and few-shot entailment models, currently effective in stance, topic, hate-speech, and event classification, shows immense potential for open, accessible, and reproducible text analysis in political science. Future research should explore expanding the capabilities of these models to new tasks (such as identifying entities, and relationships) and new document sources. While we think that these models can be immensely valuable to researchers now, we hope that this is only the first step in developing efficient, open source models tailored for specific domains. We think that there is significant room to further expand the PolNLI data set and, as a result, train better models that more widely generalize across political communication. We believe that domain adapted language models can be a public good for the research community and hope that researchers studying politics will collaborate to share data and expand the training corpus for these models.

Further, we also believe that open source LLM-based chat bots could benefit greatly from our approach of domain adaptation and entailment classification. Thus, in future work we hope to adapt and expand the PolNLI data set to make it suitable for generative language models. By doing so, it is plausible that not only could generative models smaller than Llama 8B achieve state-of-the-art classification performance, but researchers would be able to use these models for tasks encoder models are not capable of, such as synthetic data generation or summarization.

\clearpage

\bibliographystyle{chicago}

\bibliography{refs}

\clearpage

\appendix
\section{Data Sources}
\label{appendix:sources}

\begin{table}[!h]
\footnotesize
\centering
\caption{Data Sets Overview}
\label{tab:datasets}
\begin{tabular}{|p{2.5cm}|p{2.75cm}|p{1cm}|p{7cm}|}
\hline
\textbf{Data Set} & \textbf{Source} & \textbf{Task} & \textbf{Notes} \\
\hline
Multi-target Stance Detection & \citet{sobhani2017dataset} & Stance & Stance labeled tweets, each containing multiple politicians. \\
\hline
PoliBERTweet Training& \citet{kawintiranon2022polibertweet} & Stance & Tweets about Trump and Biden. \\
\hline
Polistance Affect & New Dataset & Stance & Tweets labeled for stance towards 20+ members of congress. \\
\hline
Polistance Quote Tweets & New Dataset & Stance & Quote tweets labeled for stance towards 20+ members of congress. \\
\hline
Newsletter Sentences & New Dataset & Stance & Newsletter sentences collected from DC Inbox. Labeled for stance towards 20+ members of congress\\
\hline
Political Tweets & Huggingface Hub & Stance & Tweets from senators and representatives labeled for stance on political issues. \\
\hline
ADL Heat Map Dataset & \citet{adl_heat_map} & Events & Description of antisemitic incidents with category and type labels. \\
\hline
State of the Union Speeches & \citet{jones2023policy} & Topic & Sentences from State of the Union speeches coded by topic and subtopic. \\
\hline
Democratic Party Platforms & \citet{wolbrecht2023dem} & Topic & Sentences from Democratic party platforms coded by topic and subtopic.\\
\hline
Republican Party Platforms & \citet{wolbrecht2023rep} & Topic & Sentences from Republican party platforms coded by topic and subtopic. \\
\hline
The Supreme Court Database & \citep{CiteSupremeCourtDB} and \cite{bird2009policy} & Topic & Summaries of court cases labeled by legal topic. Summaries were taken from the Comparative Agendas Project. \\
\hline
Argument Quality Ranking &  \cite{DBLP:journals/corr/abs-1911-11408} & Stance & Crowd sourced arguments for or against 71 different propositions. Subset to include only political topics. \\
\hline
Global Warming Media Stance & \citet{luo-etal-2020-detecting}& Stance & News leads labeled for if they portray global warming as a threat. \\
\hline
Claim Stance & \citet{bar-haim-etal-2017-stance} & Stance & Claims from Wikipedia across 55 topics. \\
\hline
Claim Stance & \citet{bar-haim-etal-2017-stance} & Topic & Claims from Wikipedia across 55 topics. \\
\hline
ACLED & \cite{raleigh2023political} & Events & Descriptions and headlines of violent events and political demonstrations. \\
\hline
SCAD & \citet{salehyan2012social} & Events & Summaries of conflict events in Africa and Latin America labeled by event type. \\
\hline
Measuring Hate Speech & \citet{kennedy2020constructing} & Hate & Hate speech and counter hate speech. Crowd sourced labels. \\
\hline
Anthropic Persuasion & \cite{durmus2024persuasion} & Stance & Arguments generated by Claude 2 and 3 across 75 topics. Subset to political topics. \\
\hline
Polarizing Rhetoric Tweets & \citet{ballard2023dynamics} & Hate & Tweets labeled by whether or not they use polarizing rhetoric. \\
\hline
Bill Summaries & Huggingface Hub & Topic & Bill summaries and labels from congress.gov. \\
\hline
Political or Not & New Dataset & Topic & News articles combined with samples from the other data sets. \\
\hline
\end{tabular}
\end{table}

\clearpage

\section{LLM Prompts}
\label{appendix: prompts}
\subsection{GPT-4/4o Label Validation Prompts and Arguments}
``You are a classifier that can only respond with 0 or 1. I'm going to show you a short text sample and I want you to determine if \{hypothesis\}. Here is the text:\\
\{document\}
\\
\noindent If it is true that \{hypothesis\}, return 0. If it is not true that \{hypothesis\}, return 1.
Do not explain your answer, and only return 0 or 1.''
\subsection{GPT-4o Hypothesis Augmentation Prompt}
``Write 3 sentences that are synonymous to this sentence: 
\\\{hypothesis\} \\
Format your output as a python list named `hypoths.'''

\subsection{GPT-4/4o Model Arguments}
model = ``gpt-4-1106-preview'' (for GPT-4 queries)\\
model = ``gpt-4o-2024-05-13'' (for GPT-4o queries)\\
system\_message = ``You are a text classifier and are only allowed to respond with 0 or 1''\\
max\_tokesn = 1\\
temperature = 0\\
logit\_bias = \{15:100, 16:100\}

\clearpage
\section{Training Parameters}
\label{appendix: params}
\subsection{Base Model}
lr\_scheduler\_type= ``linear''\\
group\_by\_length=False \\
learning\_rate=2e-5\\
per\_device\_train\_batch\_size=8\\
per\_device\_eval\_batch\_size=8\\
num\_train\_epochs=20\\
warmup\_ratio=0.06\\
weight\_decay=0.01\\
fp16=True\\
fp16\_full\_eval=True\\
eval\_strategy=``epoch''\\
seed=1\\
save\_strategy=``epoch''\\
dataloader\_num\_workers = 12\\

\subsection{Large Model}
lr\_scheduler\_type= ``linear''\\
group\_by\_length=False \\
learning\_rate=9e-6\\
per\_device\_train\_batch\_size=4\\
per\_device\_eval\_batch\_size=8\\
gradient\_accumulation\_steps=4\\
num\_train\_epochs=20\\
warmup\_ratio=0.06\\
weight\_decay=0.01\\
fp16=True\\
fp16\_full\_eval=True\\
eval\_strategy=``epoch''\\
seed=1\\
save\_strategy=``epoch''\\
dataloader\_num\_workers = 12\\

\end{document}